\documentclass{mva_style}
\usepackage{graphicx}
\graphicspath{{./images/}}
\usepackage{amsmath,amssymb,amsfonts}
\usepackage{textcomp}
\usepackage{xcolor}
\usepackage{comment}
\usepackage{algorithm}
\usepackage{algorithmic}
\usepackage{booktabs}
\usepackage{multirow}
\usepackage{multicol}
\usepackage{stfloats}
\usepackage{hyperref}
\usepackage{caption} 
\finalcopy %Uncomment this line for the Camera-Ready Manuscript

\begin{document}
\title{Semantic Hierarchy Preserving Deep Hashing for Large-Scale Image Retrieval}

\author{
  Ming Zhang\\
  City University of Hong Kong\\
  Hong Kong, China\\
  {\tt mzhang367-c@my.cityu.edu.hk}\\
  \and
  Xuefei Zhe\\
  Tencent AI Lab\\
  Shenzhen, China\\
  {\tt elizhe@tencent.com}\\
  \and
  Le Ou-Yang\\
  Shenzhen University\\
  Shenzhen, China\\
  {\tt leouyang@szu.edu.cn}\\
  \and
  Shifeng Chen\\
  Shenzhen Institutes of Advanced Technology, CAS\\
  Shenzhen, China\\
  {\tt shifeng.chen@siat.ac.cn}\\
  \and
  Hong Yan\\
  City University of Hong Kong\\
  Hong Kong, China\\
  {\tt h.yan@cityu.edu.hk}\\
}

\maketitle

\section*{\centering Abstract}
\textit{
   Deep hashing models have been proposed as an efficient method for large-scale similarity search. However, most existing deep hashing methods only utilize fine-level labels for training while ignoring the natural semantic hierarchy structure. This paper presents an effective method that preserves the classwise similarity of full-level semantic hierarchy for large-scale image retrieval. Experiments on two benchmark datasets show that our method helps improve the fine-level retrieval performance. Moreover, with the help of the semantic hierarchy, it can produce significantly better binary codes for hierarchical retrieval, which indicates its potential of providing more user-desired retrieval results. The codes are available at \url{https://github.com/mzhang367/hpdh.git}.
}

\section{Introduction}
The past few years have witnessed a significant improvement in the quality of content-based image retrieval (CBIR)~\cite{kulis2009fastCBIR,gong2013iterativeCBIR,li2014subCBIR} due to the emergence of deep learning. With the explosive growth of online visual data, there is an urgent need to develop more efficient deep learning models. Recently, deep hashing has been proposed as a promising method for large-scale image retrieval. It directly projects images to binary codes for approximate nearest neighbor search, which considerably reduces the storage and computation cost. 

\begin{figure}[t]
\centering
\includegraphics[width=0.35\textwidth]{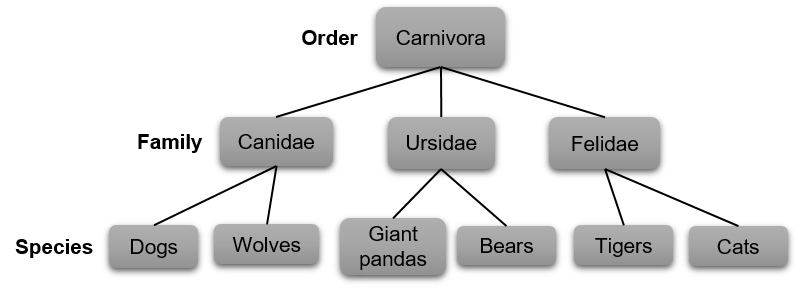}
\caption{A tree visualization of three-level hierarchy comprising six kinds of mammals. Note that we refer the Species level to fine-level and the Order level to the highest level, which is analogous to the leaf node and the root node in the hierarchy tree, respectively. 
%Here the genus level between the family level and the species level is omitted for simplicity.
}
 \label{fig:hierar_tree}
\end{figure}

In the real world, the classification and description of things often follow a hierarchy structure. One typical example is taxonomy, as shown in Fig.~\ref{fig:hierar_tree}. However, most existing deep hashing models~\cite{DPSH,DQN,li2017deep,DSHNP} only utilize single-level semantic labels for training. Their training processes are either supervised with the fine-level labels or similar/dissimilar pairs of labels that are converted from fine-level labels. With such supervision, the deep hashing model can only learn partly class similarity within the lowest hierarchy while the class similarity between the upper-level labels is not well preserved as the semantic hierarchy structure. Consequently, it hinders the deep hashing model from learning a better semantic hashing space for hierarchical retrieval. Taking Fig.~\ref{fig:hierar_tree} as an example. Without hierarchy, bears' feature embeddings are not necessarily closer to giant pandas belonging to the same super-class (Ursidae) than that to other species belonging to different super-class. 

\begin{figure}[t]
\centering
\includegraphics[width=0.45\textwidth]{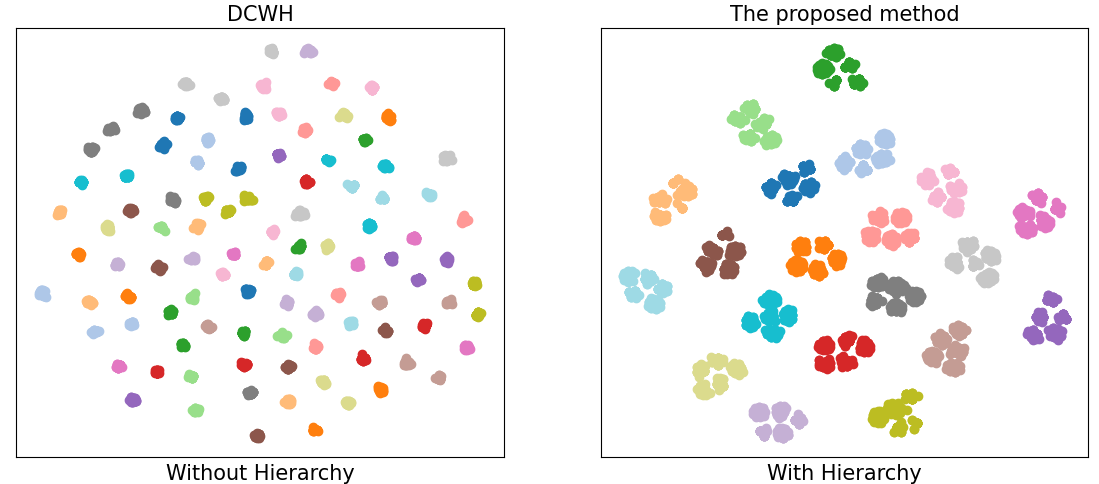}
\caption{Visualization of learned hashing codes using t-SNE~\cite{van2008visualizing}. The samples in CIFAR-100~\cite{CIFAR10} are labeled with both fine-level and coarse-level labels. We illustrate samples by their coarse-level labels. Each color indicates one coarse-level class. Note each coarse-level class contains exactly five fine-level classes.}
 \label{fig:vis_tsne}
\end{figure}
%the feature space can only guarantee that intra-class samples of bears are closer to each other than to samples from other species. Whereas, 
In this work, we propose a novel deep hashing method called Hierarchy Preserving Deep Hashing (HPDH). Fig.~\ref{fig:vis_tsne} illustrates a comparison of the fine-level labels-based Deep Class-Wise Hashing (DCWH)~\cite{zhe2019deep} and the proposed method. It is clear that the hashing codes generated by our method fall into an obvious hierarchy structure. The main contribution of HPDH can be summarized in three folds:  \begin{itemize}
  \item We propose a hierarchical loss function that directly uses class labels for hashing learning. The proposed loss function preserves intra-class compactness and inter-class separability in each hierarchy level.  
  \item To better leverage the label information's hierarchy structure, we design a simple yet efficient scheme to update the class centers per level in a periodical and recursive manner.  
  \item  Experiments on two benchmark datasets: CIFAR-100 and NABirds show that our method consistently outperforms other state-of-the-art baselines with a distinct margin on both general fine-level retrieval and hierarchical retrieval. 
\end{itemize}

\section{Related Works}
%With the availability of large labeled datasets and the growing on the capacity of computational resource
 Deep hashing methods~\cite{li2017deep,zhe2019deep, zhang2021deep, luo2020survey} have shown their great superiority to traditional hashing methods in image retrieval with the advantages of simultaneously feature representations learning and hashing learning. Most existing deep hashing methods adopt the similar/dissimilar pairs of samples for training that are constructed from class labels. 
 %Specifically, image pairs sharing at least one common class label are denoted as similar. Otherwise, they are dissimilar. 
 Following this direction, DPSH~\cite{DPSH} firstly utilizes the pairwise label information to train an end-to-end deep hashing model. HashNet~\cite{2017hashNet} defines a weighted maximum likelihood to balance the similar and dissimilar pairs within the dataset. To make the most of the semantic information, recently, several works propose directly relying on class labels for similarity supervision. SSDH~\cite{SSDH} introduces a softmax classifier to unify the classification and retrieval in a single learning process. DCWH~\cite{zhe2019deep} designs a normalized Gaussian-based loss, clustering intra-class samples to the corresponding class center.

The aforementioned deep hashing methods propose different learning metrics to conduct similarity learning. However, most of them cannot handle the multi-level semantic labels. Meanwhile, there appear more and more hierarchical labeled datasets. For example, ImageNet~\cite{ILSVRC15} is a large-scale database organized by WordNet~\cite{fellbaum2010_wordnet} hierarchy, and each node of the hierarchy is illustrated by hundreds, even thousands of images. Therefore, just considering the similarity within a single level of labels inevitably leads to a loss of the rich semantic information stored in the hierarchical data.

It has been verified that providing the hierarchy information relating to semantic labels during training can boost image retrieval performance. In~\cite{dutt2017improving}, researchers prove that the image classification performance can be improved by combining the coarse and fine-level labels. A similar idea is shared in~\cite{ma2018hierarchical}, where a hierarchical training strategy is applied to handle the face recognition task. Recently, SHDH~\cite{wang2017Hierarchical} is the first deep hashing work that tackles the hierarchy similarity by weighting the Hamming distance at each level. However, SHDH applies a pairwise labels relation, which is inferior to class-wise labels-based methods~\cite{SSDH,zhe2019deep, zhang2021improved}. Motivated by these issues, we introduce the hierarchy -preserving method HPDH based on the class-wise label information in each hierarchy, suitable for learning more discriminative binary codes.

\section{Proposed Approach}
Suppose a hierarchical labeled dataset, and each image $x_i$ is annotated with a $K$-level semantic label, denoted with a $K$-dimensional vector $y_{i}$. The label vector consists of the class labels from the lowest hierarchy level to the highest hierarchy level in a tree-like structure. For example, a dolphin image is labeled with $y=\{$Dolphin, Aquatic Mammals$\}$ in a two-level hierarchy. We aim to apply deep convolutional neural network (CNN) with learnable parameters $\Theta$ to project images to a particular Hamming space. In this space, for any hierarchy level, the Hamming distance between intra-class samples to the corresponding class center is smaller than that to other class centers.

We propose a multi-level normalized Gaussian model to keep a hierarchical semantic structure in the Hamming space. %Consequently, the trained models can only preserve the similarity relation between the same lowest class while the similarities within all the parent classes are out-of-order. To keep a semantic hierarchical structure in the Hamming space, we propose a multi-level normalized Gaussian model.
Denote the hashing output as $r_i=f(x_i, \Theta)$ and $r_i\in \{-1,1\}^L$, where $L$ is the binary code length. The proposed objective function is formulated as following:
\begin{gather} \label{eq:obj}
\min_{\Theta, M}\mathcal{L} = -\sum_{i=1}^N \sum_{k=1}^K \log\frac{\exp\{-\frac{1}{2 \sigma_k^2} d(r_i,{\mu_{ky_{ik}}})\}}{\sum_{j=1}^{C_k} \exp\{-\frac{1}{2 \sigma_k^2} d(r_i ,{\mu}_{kj}) \}}\\
s.t. \quad r_i = f(x_i,\Theta)\in \{-1,1\}^L, \quad \mu_{kj} \in \{-1,1\}^L \nonumber
\end{gather}
where $M=\{\mu_k\}_{k=1}^K$ and $\mu_{k}=\{\mu_{kj}\}_{j=1}^{C_k}$. Here, $\mu_{kj}$ represents the $j$-th class centers in the $k$-th level of the semantic hierarchy, $C_k$ is the total number of classes at the $k$-th level and $\sigma_k$ is a parameter to control the intra-class variance at the level $k$. $d(\cdot,\cdot)$ is the Hamming distance function. To solve this discrete optimization problem, we follow the optimization strategy in DCWH~\cite{zhe2019deep}. We first relax the $r_i$ to $[-\alpha,\alpha]$ where $\alpha$ is empirically set to $1.1$ in~\cite{zhe2019deep}. Then, the original distance $d(\cdot,\cdot)$ can be replaced with a Euclidean distance. And the loss function becomes the following:
\begin{equation}
\label{eq:loss}
\begin{split}
    \mathcal{L}= &-\sum_{i=1}^N \sum_{k=1}^K \log\frac{\exp\{-\frac{1}{2 \sigma_k^2 } \|r_i-{\mu}_{ky_{ik}}\|\}}{\sum_{j=1}^{C_k} \exp\{-\frac{1}{2 \sigma_k^2} \|r_i -{\mu}_{kj}\| \}} \\
    &+\eta_1\{ReLU(-\alpha-r_n) +ReLU(r_n-\alpha) \}
\end{split}
\end{equation}
where $\eta_1$ is the regularization weight. ${ReLU}$ is the rectified linear unit defined as $ReLU(x) = \max(0,x)$. The above loss function is differentiable and the classical back-propagation can be applied to optimize the network parameters $\Theta$.

We update the fine-level class centers $\{\mu_{1j}\}$ with the training data as following:
\begin{equation}
    \label{eq:up_1}
    \mu_{1j} = \frac{1}{N_{1j}}\sum_{n=1}^{N_{1j}}~f(x_n,\Theta)
\end{equation}
where $N_{1j}$ is the number of images that belong to the $j$-th class in the lowest hierarchy level in the whole dataset. Based on $\{\mu_{1j}\}$, the upper-level class centers $\{\mu_{kj}\}_{k=2}^K$ can be calculated from their own child level class centers as following:
\begin{equation}
\label{eq:up_2}
     \mu_{kj} = \frac{1}{C_{kj}}\sum_{c=1}^{C_{kj}}~\mu_{(k-1),c}
\end{equation}
where $C_{kj}$ is the number of level-($k$-1) classes with the same parent class, i.e., $j$-th class at level $k$. By such a recursive calculation, not only can we save the computation cost but also eliminate the influence of imbalanced training data between classes. The hyper parameters $\{\sigma_k^2\}$ are chosen to satisfy the criterion $\sigma^2_{k-1} \leq \sigma^2_{k}$ so that the variances within parent classes are larger than the child ones. Finally, we introduce a quantization term following~\cite{li2017deep} to encourage the relaxed real-valued hashing outputs to be binary. The finalized objective function is shown as:
\begin{equation}\label{eq:fin}
 \min_{\Theta,M}\mathcal{J} = \mathcal{L} + \eta_2 \sum_{i=1}^{N}\left\|b_{i}-r_{i}\right\|_{2}^{2}
\end{equation}
where $b_i=\mathbf{sgn}(r_i)$ and $\eta_2$ is a hyper parameter controlling the weight of the quantization term. The whole training procedure is summarized in Algorithm~\ref{alg:1}.
\begin{algorithm}
\caption{The training procedure of HPDH}\label{alg:1}
\begin{algorithmic}

\STATE\textbf{Initialize} CNN parameters $\Theta$, class centers $M$
\REPEAT
	\STATE 1. Compute features $r_i=f(x_i,\Theta)$;
    \STATE 2. Update fine-level centers $\{\mu_{1j}\} $ by Eq.~(\ref{eq:up_1});
    \STATE 3. Update upper level centers by Eq.~(\ref{eq:up_2});
    \STATE 4. Compute the loss $\mathcal{J}$ according to Eq.~(\ref{eq:fin});
    \STATE 5. Calculate derivatives of $\mathcal{J}$ w.r.t. $r_i$ and update $\Theta$ by back propagation
\UNTIL{Converge}
\end{algorithmic}
\end{algorithm}

\section{Experiments}
\subsection{Datasets}
We conduct experiments on two hierarchical datasets including CIFAR-100~\cite{CIFAR10} and NABirds~\cite{NAbirds}. CIFAR-100 is a dataset collecting $60,000$ tinny images with a two-level semantic hierarchy, i.e., $K=2$. Specifically, the total $100$ fine-level classes are grouped into $20$ coarse categories, and each coarse category contains exactly $5$ fine-level classes. We follow the official split with $50,000$ images for training and $10,000$ images for testing. NABirds~\cite{NAbirds} dataset comprises more than $48,000$ images from $555$ visual categories of North America birds species. These categories are organized taxonomically into a four-level hierarchy (excluding the root node ``bird'' that all images belong to, which does not provide any information gain). We use the official split that has $23,929$ images and $24,633$ images for training and testing, respectively. 
%in which each fine-level class has $500$ images for training and $100$ images for testing. \footnote{\url{https://www.cs.toronto.edu/~kriz/cifar.html}} and NABirds \footnote{\url{https://dl.allaboutbirds.org/nabirds}}
In the training procedure, we directly resize the original images in two datasets to $240\times240$ and then randomly crop them to $224\times224$ as inputs of the network. For both datasets, the training set serves as the database, and the testing images are used as queries during the testing phase. 

\subsection{Setups and Evaluation Metrics}
We compare our method with a series of pairwise and triplet labels-based deep hashing methods, including DPSH~\cite{DPSH}, DTSH~\cite{DTSH}, SHDH~\cite{wang2017Hierarchical}, and classwise labels-based methods, including DCWH~\cite{zhe2019deep}, IDCWH~\cite{zhang2021improved}, and CSQ~\cite{yuan2020central}. For a fair comparison, we apply ResNet-50~\cite{he2016resNet} pre-trained on the ImageNet~\cite{ILSVRC15} in all the methods. For our HPDH, we fine-tune the backbone with a newly added fully-connected hashing layer.
%The FCH layer is initialized by the Xavier method with the magnitude of $1$. 
We train the whole network for 150 epochs using mini-batch stochastic gradient descent (SGD) with momentum 0.9 and weight decay 5e-4. The initial learning rate is set to 5e-3 and decayed by 0.1 every 50 epochs. The hyper parameters are fixed to $\eta_1=10$ and $\eta_2=0.1$. We set $\{\sigma_k\} = \{1, 2\}$ for CIFAR-100 and $\{1, 1.5, 2, 4\}$ for NABirds by cross-validation. All the experiments are run on two Nvidia RTX-2080 GPU cards with PyTorch.

We present results under both fine-level and hierarchical-level evaluation metrics. The fine-level retrieval performance is evaluated with the mean average precision (mAP@all). While the hierarchical retrieval performance is reported by mean average Hierarchical Precision (mAHP)~\cite{wang2017Hierarchical,barz2019hierarchy} and the normalized Discounted Cumulative Gain (nDCG)~\cite{jarvelin2002cumulated}. Specifically, we adopt mAHP@2,500 for CIFAR-100 which contains exactly $2,500$ images for each coarse category, while mAHP@250 for NABirds dataset following~\cite{barz2019hierarchy}. We provide results nDCG@100 for easy comparison with~\cite{wang2017Hierarchical}. We refer readers to~\cite{deng2011hierarchical} for more details about hierarchical precision (HP).

\begin{comment}
Considering $x_q$ is a query image with label vector $y_q$, the ordered retrieval results are denoted as $\{(x_1,y_1),(x_2,y_2),\dots,(x_N,y_N)\}$. The hierarchical precision (HP)~\cite{deng2011hierarchical} at N-th returned images is defined as 
\begin{equation}
    HP@N = \frac{\sum_{i=1}^{n}Sim(y_q,y_i)}{max_o \sum_{i=1}^{n}Sim(y_q,y_{o_i})}
\label{eq:hp}
\end{equation}
where $Sim(\cdot, \cdot)$ is a score function for measuring the similarity between two label vectors based on the category hierarchy. The denominator in Eq.~(\ref{eq:hp}) represents the largest summation that can be achieved by the best $N$ database items. We refer readers to \cite{deng2011hierarchical} for more details about HP.
\end{comment}

\begin{table*}[t]
\caption{Results on CIFAR-100 dataset}
\label{tab:c100}
\centering
\small
\begin{tabular}{cccccccccccc}
\hline
\multirow{2}{*}{Methods} &\multicolumn{3}{c}{mAP@all} &  &\multicolumn{3}{c}{mAHP@2.5k} & &\multicolumn{3}{c}{nDCG@100}\\ \cline{2-4} \cline{6-8} \cline{10-12}
                         & 32-bit  & 48-bit  & 64-bit &  & 32-bit   & 48-bit  & 64-bit & &32-bit &48-bit &64-bit \\ \hline
DPSH                     &0.2861  &0.3571  &0.3952  &  &0.4715 &0.5048   &0.5212 &  &0.5409  &0.6089 &0.6442\\
DTSH                     &0.6950  &0.7269  &0.7366  &  &0.6744 &0.7019   &0.7051 &  &0.7468  &0.7625 &0.7820\\
SHDH                    &-  &-  &-  &  &-  &-  &- &  &0.6141  &0.6281 &0.6406\\
DCWH                     &0.7680  &0.8023  &0.8178  &  &0.6598 &0.6608   &0.6698 &  &0.7894 &0.8306 &0.8439\\
CSQ                     &0.7991  &0.8032  &0.8093  &  &0.5660 &0.5677   &0.5792 &  &0.8348  &0.8376 &0.8387 \\
HPDH        & \textbf{0.8292}  & \textbf{0.8347}  & \textbf{0.8534}  &  & \textbf{0.8802}   & \textbf{0.8846}  & \textbf{0.8974}  &  & \textbf{0.8520}   & \textbf{0.8558}  & \textbf{0.8718}\\ \hline
\end{tabular}
\end{table*}

\begin{table*}[t]
\caption{Results on NABirds dataset}
\label{tab:birds}
\centering
\small
\begin{tabular}{cccccccccccc}
\hline
\multirow{2}{*}{Methods} &\multicolumn{3}{c}{mAP@all} &  &\multicolumn{3}{c}{mAHP@250} & &\multicolumn{3}{c}{nDCG@100}\\ \cline{2-4} \cline{6-8} \cline{10-12}
                         & 32-bit  & 48-bit  & 64-bit &  & 32-bit   & 48-bit  & 64-bit & &32-bit &48-bit &64-bit \\ \hline
DTSH                     &0.3511  &0.3614  &0.3764  &  &0.6514 &0.6566   &0.6634 &  &0.5702  &0.5798 &0.5902\\
DCWH                     &0.4132  &0.4882  &0.5132  &  &0.5715 &0.6400   &0.6637 &  &0.5954 &0.6551 &0.6754 \\
CSQ                     &0.4419  &0.4733  &0.5121  &  &0.4557 &0.4998   &0.5390 &  &0.5445  &0.5981 &0.6396\\
IDCWH                    &0.6811  &0.7158  &0.7249  &  &0.5857 &0.6389  &0.6843 &  &0.7228 &0.7602 &0.7878\\
HPDH        & \textbf{0.7014}  & \textbf{0.7372}  & \textbf{0.7366}  &  & \textbf{0.8512}   & \textbf{0.8661}  & \textbf{0.8560}  &  & \textbf{0.7987}   & \textbf{0.8160}  & \textbf{0.8040}\\ \hline
\end{tabular}
\end{table*}

\subsection{Results and Analysis}

\begin{figure}[htbp]
\centering
\includegraphics[width=0.45\textwidth]{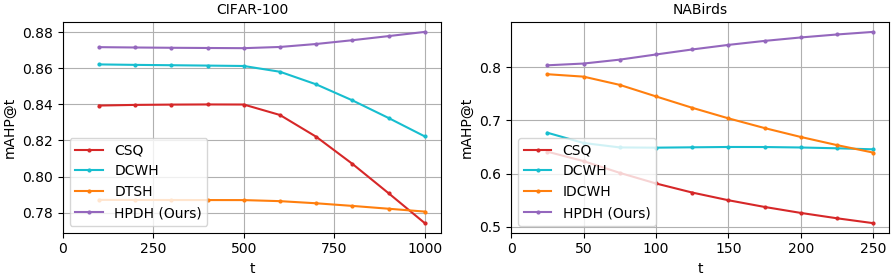}
\caption{The comparison on mAHP results w.r.t. different cutoff points $t$ on two datasets.}
 \label{fig:plot_hp}
\end{figure}

We present results on CIFAR-100 in Table~\ref{tab:c100}. We can find, for general retrieval performance, the proposed HPDH performs better than two state-of-the-art, i.e., DCWH and CSQ. The mAP of HPDH is 3.0$\%$ and 3.6$\%$ higher than that of the second place at 32-bit and 64-bit, respectively. When it comes to hierarchical retrieval, our method surpasses the previous best one with a large margin: it obtains 88.74$\%$ mAHP scores on average, which is 19.4$\%$ higher than DTSH's 69.38$\%$. Note the distinct performance drop from mAP to mAHP for DCWH and CSQ. In contrast, our HPDH utilizing the all-level hierarchy labels performs even better on mAHP than mAP. It indicates that the semantic hierarchical Hamming space is hard to learn if only use the fine-level class labels. From the comparison with SHDH, which also utilizes hierarchy labels but pairwise label similarity, our method significantly outperforms SHDH by 23.2$\%$ under nDCG metric in average. The performance on NABirds dataset is presented in Table~\ref{tab:birds}. We add the comparison with latest IDCWH. 
%Here we no longer show the results on DPSH since the performance of DPSH significant deteriorates with the increase on the number of classes. 
We can observe that the proposed HPDH performs the best on NABirds in general retrieval and hierarchical retrieval tasks. Specifically, it achieves 73.72$\%$ under mAP and 81.60$\%$ under nDCG at 48-bit, with a superiority of 2.1$\%$ and 5.6$\%$ to that of IDCWH. While for mAHP, HPDH obtains the best result 86.61$\%$ at 48-bit, surpasses the second place DTSH by a nearly $21\%$ margin. 

In Fig.~\ref{fig:plot_hp}, we plot the mAHP@$t$ results w.r.t. different cutoff points $t$ in CIFAR-100 and NABirds datasets, respectively. From Fig.~\ref{fig:plot_hp}, one phenomenon worth noting is, there is a declining trend of mAHP with the increase on $t$ for all the compared methods, while only our method improves mAHP with the growth on $t$. Specifically, all the methods have a turning point at $t=500$ in the curves of CIFAR-100. The reason is there are exactly 500 images per fine-level class, and it is no longer sufficient to retrieve the images with exactly the same labels after this point. However, the proposed HPDH, which benefits from learning the hierarchical similarity information, is the only method capable of retrieving similar images with the same parent class at later positions. Thus, it proves the effectiveness of the proposed method for discriminative hierarchical retrieval.

\section{Conclusion}
We propose a novel deep hashing model towards fully utilizing the hierarchy structure of the semantic information. Our method is based on a multi-level Gaussian loss function, and it takes the advantages of class-level similarity learning and full-level hierarchy labels in training. Experiments on two hierarchical datasets show that our method not only helps improve the fine-level retrieval performance but also results in state-of-the-art results regarding hierarchical retrieval.

\bibliographystyle{IEEEtran}
\bibliography{reference.bib}

\end{document}